%% file: main.tex
\documentclass[11pt,a4paper]{article}
\usepackage[hyperref]{emnlp-ijcnlp-2019}
\usepackage{times}
\usepackage{latexsym}
\usepackage{graphicx}
\usepackage{todonotes}
\usepackage{amsmath}

\usepackage{graphicx}
\usepackage{booktabs}
\usepackage{multirow}
\usepackage{rotating}
\usepackage{amsmath}
\usepackage{dsfont}

\usepackage[toc,page]{appendix}
\usepackage{xspace}
\usepackage{array}
\usepackage{enumitem}
\usepackage{pifont}
\usepackage{url}

\aclfinalcopy 

\newcommand\dataset{\textsc{Quoref}\xspace}
\newcommand{\secref}[1]{\S\ref{#1}}

\title{\dataset{}: A Reading Comprehension Dataset with\\ Questions Requiring Coreferential Reasoning}

\newcommand{\andd}{\hspace{1.5em}}
\newcommand{\aii}{$^\heartsuit$}
\newcommand{\uw}{$^\clubsuit$}

\author{Pradeep Dasigi\aii \andd Nelson F. Liu\aii\uw \andd  Ana Marasovi\'{c}\aii \\
\textbf{Noah A. Smith}\aii\uw \andd \textbf{Matt Gardner}\aii \\
  \aii Allen Institute for Artificial Intelligence \\
    \uw Paul G. Allen School of Computer Science \& Engineering, University of Washington \\
      {\hypersetup{urlcolor=black}
        {\tt \{\href{mailto:pradeepd@allenai.org}{pradeepd},\href{mailto:anam@allenai.org}{anam},\href{mailto:mattg@allenai.org}{mattg}\}@allenai.org},
	  {\tt \{\href{mailto:nfliu@cs.washington.edu}{nfliu,\href{mailto:nasmith@cs.washington.edu}{nasmith}\}@cs.washington.edu}}
	    }
	      }

		\date{}

\begin{document}
		\maketitle
		\begin{abstract}
			Machine comprehension of texts longer than a single sentence often requires coreference resolution. However, most current reading comprehension benchmarks do not contain complex coreferential phenomena and hence fail to evaluate the ability of models to resolve coreference.
			We present a new crowdsourced dataset containing more than 24K span-selection questions that require resolving coreference among entities in over 4.7K English paragraphs from Wikipedia. Obtaining questions focused on such phenomena is challenging, because it is hard to avoid lexical cues that shortcut complex reasoning.
			We deal with this issue by using a strong baseline model as an adversary in the crowdsourcing loop, which helps crowdworkers avoid writing questions with exploitable surface cues. We show that state-of-the-art reading comprehension models perform significantly worse than humans on this benchmark---the best model performance is 70.5 $F_1$, while the estimated human performance is 93.4 $F_1$.
		\end{abstract}

		\section{Introduction}
		\begin{figure}[ht]
			     \centering
			          \includegraphics[width=3.0in]{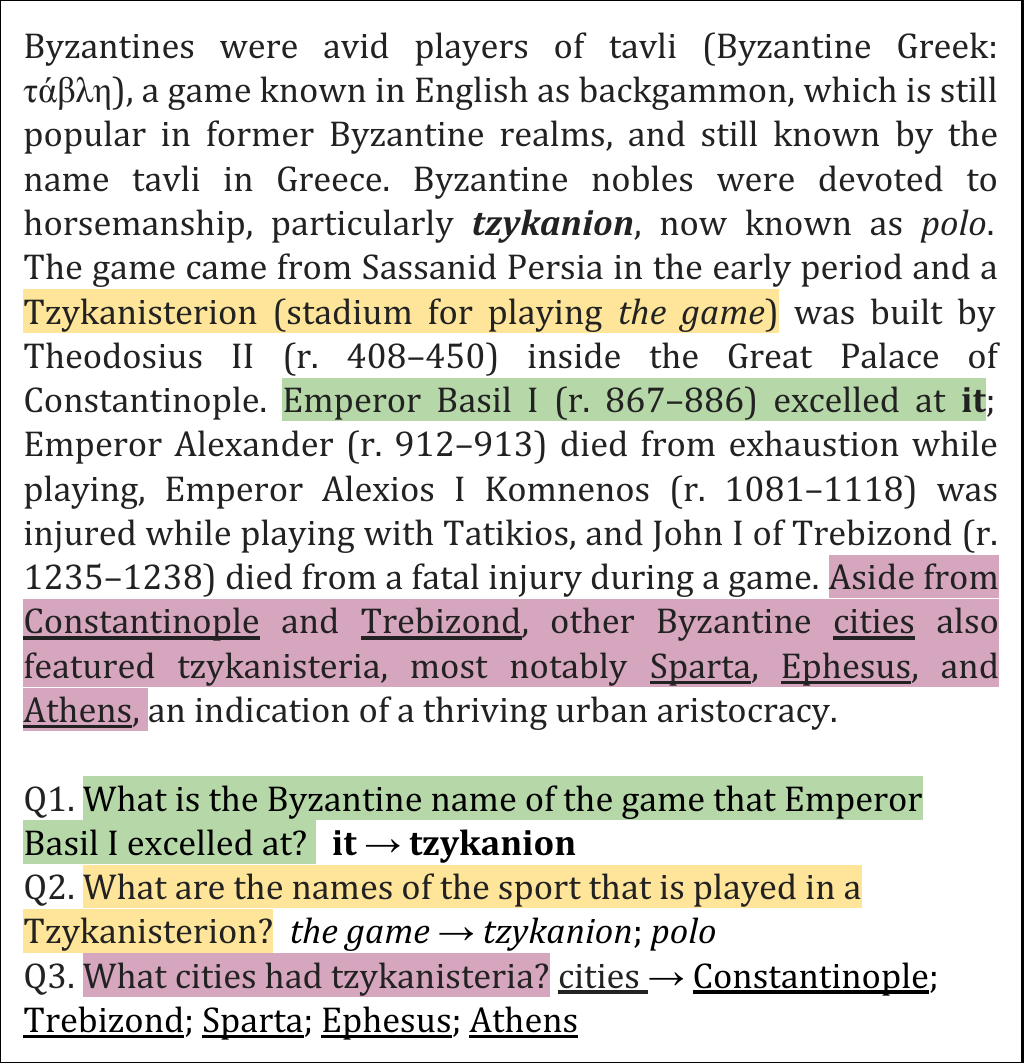}
				       \caption{Example paragraph and questions from the dataset. Highlighted text in paragraphs is where the questions with matching highlights are anchored. Next to the questions are the relevant coreferent mentions from the paragraph. They are bolded for the first question, italicized for the second, and underlined for the third in the paragraph.}
				            \label{fig:quoref_example}
					     \end{figure}

					     Paragraphs and other longer texts typically make multiple references to the same entities. Tracking these references and resolving coreference is essential for full machine comprehension of these texts.  Significant progress has recently been made in reading comprehension research, due to large crowdsourced datasets~\cite[\textit{inter alia}]{rajpurkar2016squad,bajaj2016ms,Joshi2017TriviaQAAL,Kwiatkowski2019NQ}.  However, these datasets focus largely on understanding local predicate-argument structure, with  very few questions requiring long-distance entity tracking.  Obtaining such questions is hard for two reasons: (1) teaching crowdworkers about coreference is challenging, with even experts disagreeing on its nuances~\citep{pradhan2007unrestricted,Versley2008VaguenessAR, recasens2011identity, poesio-etal-2018-anaphora},
					     and (2) even if we can get crowdworkers to target coreference phenomena in their questions, these questions may contain giveaways that let models arrive at the correct answer without  performing the desired reasoning (see~\secref{sec:analysis} for examples).


					     We introduce a new dataset, \dataset{},\footnote{Links to dataset, code, models, and leaderboard available at \url{https://allennlp.org/quoref}.} that contains questions requiring coreferential reasoning (see examples in Figure~\ref{fig:quoref_example}). The questions are derived from paragraphs taken from a diverse set of English Wikipedia articles and are collected using an annotation process (\secref{sec:quoref_annotation}) that deals with the aforementioned issues in the following ways:
					     First, we devise a set of instructions that gets workers to find anaphoric expressions and their referents, asking questions that connect two mentions in a paragraph.  These questions mostly revolve around traditional notions of coreference (Figure~\ref{fig:quoref_example} Q1), but they can also involve referential phenomena that are more nebulous (Figure~\ref{fig:quoref_example} Q3).
					     Second, inspired by~\citet{dua2019drop}, we disallow questions that can be answered by an adversary model (uncased base BERT, \citealp{devlin2019bert}, trained on SQuAD 1.1, \citealp{rajpurkar2016squad}) running in the background as the workers write questions. This adversary is not particularly skilled at answering questions requiring coreference, but can follow obvious lexical cues---it thus helps workers avoid writing questions that shortcut coreferential reasoning.

					     \dataset{} contains more than 24K questions whose answers are spans or sets of spans in 4.7K paragraphs from English Wikipedia that can be arrived at by resolving coreference in those paragraphs. We manually analyze a sample of the dataset (\secref{sec:analysis}) and find that 78\% of the questions cannot be answered without resolving coreference. We also show (\secref{sec:baselines}) that the best system performance is
					      70.5\% $F_1$, while the estimated human performance is 93.4\%. These findings indicate that this dataset
					      is an appropriate benchmark for coreference-aware reading comprehension. 

					      \section{Dataset Construction}\label{sec:quoref_annotation}
					      \paragraph{Collecting paragraphs}
					      We scraped paragraphs from Wikipedia pages about English movies, art and architecture, geography, history, and music. For movies, we followed the list of English language films,\footnote{\url{https://en.wikipedia.org/wiki/Category:English-language_films}} and extracted plot summaries that are at least 40 tokens, and for the remaining categories, we followed the lists of featured articles.\footnote{\url{https://en.wikipedia.org/wiki/Wikipedia:Featured_articles}} Since movie plot summaries usually mention many characters, it was easier to find hard \dataset{} questions for them, and we sampled about 40\% of the paragraphs from this category.

					      \input{tables/dataset_stats.tex}

					      \paragraph{Crowdsourcing setup}
					      We crowdsourced questions about these paragraphs on Mechanical Turk. We asked workers to find two or more co-referring spans in the paragraph, and to write questions such that answering them would require the knowledge that those spans are coreferential. We did not ask them to explicitly mark the co-referring spans. 
					      Workers were asked to write questions for a random sample of paragraphs from our pool, and we showed them examples of good and bad questions in the instructions (see Appendix~\ref{sec:crowdsourcing_logistics}). For each question, the workers were also required to select one or more spans in the corresponding paragraph as the answer, and these spans are not required to be same as the coreferential spans that triggered the questions.\footnote{For example, the last question in Table~\ref{tab:dataset_reasoning} is about the coreference of \{\textit{she}, \textit{Fania}, \textit{his mother}\}, but none of these mentions is the answer.} We used an uncased base BERT QA model \citep{devlin2019bert} trained on SQuAD 1.1~\citep{rajpurkar2016squad} as an adversary running in the background that attempted to answer the questions written by workers in real time, and the workers were able to submit their questions only if their answer did not match the adversary's prediction.\footnote{Among models with acceptably low latency, we qualitatively found uncased base BERT to be the most effective.} Appendix~\ref{sec:crowdsourcing_logistics} further details the logistics of the crowdsourcing tasks. Some basic statistics of the resulting dataset can be seen in Table~\ref{tab:dataset_stats}.

					      \input{tables/analysis_examples.tex}
					      \section{Semantic Phenomena in \dataset{}}\label{sec:analysis}
					      To better understand the phenomena present in \dataset{}, we manually analyzed a random sample of 100 paragraph-question pairs. The following are some empirical observations.

					      \paragraph{Requirement of coreference resolution}
					      We found that 78\% of the manually analyzed questions cannot be answered without coreference resolution. The remaining 22\% involve some form of coreference, but do not require it to be resolved for answering them. Examples include a paragraph that mentions only one city, ``\textit{Bristol}'', and a sentence that says ``\textit{the city was bombed}''. The associated question, \textit{Which city was bombed?}, does not really require coreference resolution from a model that can identify city names, making the content in the question after \textit{Which city} unnecessary. 

					      \paragraph{Types of coreferential reasoning} Questions in \dataset{} require resolving pronominal and nominal mentions of entities. Table~\ref{tab:dataset_reasoning} shows percentages and examples of analyzed questions that fall into these two categories.  These are not disjoint sets, since we found that 32\% of the questions require both (row 3). We also found that 10\% require some form of commonsense reasoning (row 4).

					      \section{Baseline Model Performance on \dataset{}}\label{sec:baselines}
					      \input{3_baselines.tex}

					      \section{Related Work}
					      \paragraph{Traditional coreference datasets} Unlike traditional coreference annotations in datasets like those of \citet{pradhan2007unrestricted}, \citet{ghaddar2016wikicoref}, \citet{chen2018preco} and \citet{poesio-etal-2018-anaphora}, which aim to obtain complete coreference clusters, our questions require understanding coreference between only a few spans. While this means that the notion of coreference captured by our dataset is less comprehensive, it is also less conservative and allows questions about coreference relations that are not marked in OntoNotes annotations. Since the notion is not as strict, it does not require linguistic expertise from annotators, making it more amenable to crowdsourcing. \citet{guha2015removing} present the limitations of annotating coreference in newswire texts alone, and like us, built a non-newswire coreference resolution dataset focusing on Quiz Bowl questions. There is some other recent work~\citep{poesio2019crowdsourced,aralikatte2019model} in crowdsourcing coreference judgments that relies on a relaxed notion of coreference as well.

					      \paragraph{Reading comprehension datasets} There are many reading comprehension datasets~\cite[\emph{inter alia}]{Richardson2013MCTestAC,rajpurkar2016squad,Kwiatkowski2019NQ,dua2019drop}. Most of these datasets principally require understanding local predicate-argument structure in a paragraph of text. \dataset{} also requires understanding local predicate-argument structure, but makes the reading task harder by explicitly querying anaphoric references, requiring a system to track entities throughout the discourse.

					      \section{Conclusion}
					      We present \dataset{}, a focused reading comprehension benchmark that evaluates the ability of models to resolve coreference. We crowdsourced questions over paragraphs from Wikipedia, and manual analysis confirmed that most cannot be answered without coreference resolution. We show that current state-of-the-art reading comprehension models perform significantly worse than humans. Both these findings provide evidence that \dataset{} is an appropriate benchmark for coreference-aware reading comprehension.

					      \section*{Acknowledgments}

					      We thank the anonymous reviewers for the useful discussion. Thanks to HuggingFace for releasing \texttt{pytorch-transformers}, and to Dheeru Dua for sharing with us the crowdsourcing setup used for DROP.

					      \bibliography{main}
					      \bibliographystyle{acl_natbib}

					      \newpage
					      \clearpage

					      \begin{appendices}
						      \input{appendix.tex}
					      \end{appendices}
				     \end{document}

%% file: tables/dataset_stats.tex
\begin{table}[!t]
	\centering
	\resizebox{\columnwidth}{!}{
		\begin{tabular}{lrrr}
			\toprule
			& \textbf{Train} & \textbf{Dev.} & \textbf{Test} \\
			\midrule
			Number of questions	& 19399	& 2418 & 2537 \\
			Number of paragraphs & 3771 & 454 & 477 \\
			\midrule
			Avg. paragraph len (tokens) & 384$\pm$105 & 381$\pm$101 & 385$\pm$103 \\
			Avg. question len (tokens) & 17$\pm$6 & 17$\pm$6 & 17$\pm$6 \\
			\midrule
			Paragraph vocabulary size & 57648 & 18226 & 18885 \\
			Question vocabulary size & 19803 & 5579 & 5624 \\
			\midrule
			\% of multi-span answers & 10.2 & 9.1 & 9.7 \\
			\bottomrule
		\end{tabular}
		}
		\caption{Key statistics of \dataset splits.}
		\label{tab:dataset_stats}
\end{table}

%% file: tables/analysis_examples.tex
\begin{table*}[!ht]
\centering
\resizebox{\textwidth}{!}{
\begin{tabular}{>{\raggedright}p{3cm}p{8cm}p{4cm}p{2cm}}
\toprule
\textbf{Reasoning} & \textbf{Paragraph Snippet} & \textbf{Question} & \textbf{Answer} \\
\midrule
\textbf{Pronominal resolution (69\%)} &
Anna and Declan eventually make their way on foot to a roadside pub, where they discover {\bf the three van thieves} going through Anna's luggage. {\bf \color{teal}Declan fights \emph{them}}, displaying unexpected strength for a man of his size, and retrieves Anna's bag. &
Who does {\bf \color{teal}Declan get into a fight} with? &
the three van thieves
\\
\midrule
\textbf{Nominal resolution (54\%)} &
 Later, {\bf Hippolyta} was granted a daughter, Princess Diana, $\ldots$ {\bf \color{violet}Diana defies \emph{her mother}} and $\ldots$ &
 What is the name of the person who is {\bf \color{violet} defied by her daughter}? &
 Hippolyta
 \\
 \midrule
 \textbf{Multiple resolutions (32\%)} &
 The now upbeat collective keep {\bf the toucan}, nicknaming it ``{\bf Amigo}'' $\ldots$ When authorities show up to catch {\bf the bird}, Pete and Liz spirit {\bf him} away by {\bf \color{brown} Liz hiding \emph{him} in her dress} $\ldots$ &
 What is the name of the character who {\bf \color{brown} hides in Liz's dress}? &
 Amigo
 \\
 \midrule
\textbf{Commonsense reasoning (10\%)} &
 Amos reflects back on his early childhood $\ldots$ with {\bf \color{blue} his mother \emph{Fania} and father Arieh}. $\ldots$ One of {\bf his mother}'s friends is killed while hanging up laundry during the war. $\ldots$ {\bf Fania} falls into a depression$\ldots$ {\bf she} $\ldots$ goes to $\ldots$ Tel Aviv, where {\bf she} kills {\bf herself} by overdose $\ldots$ &
 How does {\bf \color{blue}Arieh's wife} die? &
 kills herself by overdose \\
\bottomrule
\end{tabular}
}
\caption{Phenomena in \dataset. Note that the first two classes are not disjoint. In the final example, the paragraph does not explicitly say that \textit{Fania} is \textit{Arieh}'s wife.}
\label{tab:dataset_reasoning}
\end{table*}

%% file: 3_baselines.tex
We evaluated two classes of baseline models on \dataset: state-of-the art reading comprehension models that predict single spans (\S\ref{sec:single_span_rc}) and heuristic baselines to look for annotation artifacts (\S\ref{sec:heuristic_baselines}).

We use two evaluation metrics to compare model performance: exact match (EM), and a (macro-averaged) $F_1$ score that measures overlap between a bag-of-words representation of the gold and predicted answers. We use the same implementation of EM as SQuAD, and we employ the $F_1$ metric used for DROP \citep{dua2019drop}.
See Appendix~\ref{sec:experimental_setup_details} for model training hyperparameters and other details.

\input{tables/baselines_results.tex}

\subsection{Reading Comprehension Models}\label{sec:single_span_rc}

We test four single-span (SQuAD-style) reading comprehension models: (1) \textbf{QANet} \citep{Yu2018QANetCL}, currently the best-performing published model on SQuAD 1.1 without data augmentation or pretraining; (2) \textbf{QANet + BERT}, which enhances the QANet model by concatenating frozen BERT representations to the original input embeddings; (3) \textbf{BERT QA} \citep{devlin2019bert}, the adversarial baseline used in data construction, and (4) \textbf{XLNet QA} \cite{Yang2019XLNetGA}, another large pretrained language model based on the Transformer architecture \cite{Vaswani2017AttentionIA} that outperforms BERT QA on reading comprehension benchmarks SQuAD and RACE \cite{lai-etal-2017-race}.

We use the AllenNLP \citep{Gardner2018AllenNLP} implementation of QANet modified to use the marginal objective proposed by \citet{Clark2018SimpleAE} and \texttt{pytorch-transformers}\footnote{\url{https://github.com/huggingface/pytorch-transformers}} implementation of base BERT QA\footnote{The large BERT model does not fit in the available GPU memory.} and base XLNet QA. BERT is pretrained on English Wikipedia and BookCorpus \cite{Zhu2015AligningBA} (3.87B wordpieces, 13GB of plain text) and XLNet additionally on  Giga5 \citep{Napoles2012AnnotatedG}, ClueWeb 2012-B (extended from \citealp{Smucker2009ExperimentsWC}), and Common Crawl\footnote{\url{https://commoncrawl.org/}} (32.89B wordpieces, 78GB of plain text).

\subsection{Heuristic Baselines}\label{sec:heuristic_baselines}

In light of recent work exposing predictive artifacts in crowdsourced NLP datasets \citep[\textit{inter alia}]{gururangan2018annotation,Kaushik2018HowMR}, we estimate the effect of predictive artifacts by training BERT QA and XLNet QA to predict a single start and end index given only the passage as input (\textbf{passage-only}).

\subsection {Results}

Table~\ref{tab:baselines_results} presents the performance of all baseline models on \dataset. 

The best performing model is XLNet QA, which reaches an $F_1$ score of $70.5$ in the test set. However, it is still more than $20$ $F_1$ points below human performance.\footnote{Human performance was estimated from the authors' answers of 400 questions from the test set, scored with the same metric used for systems.}

BERT QA trained on \dataset{} under-performs XLNet QA, but still gets a decent $F_1$ score of $66.4$. Note that BERT QA trained on SQuAD would have achieved an $F_1$ score of 0, since our dataset was constructed with that model as the adversary. The extent to which BERT QA does well on \dataset{} might indicate its capacity for coreferential reasoning that was not exploited when it was trained on SQuAD (for a detailed discussion of this phenomenon, see~\citealp{liu2019inoculation}). Our analysis of model errors in~\secref{sec:error_analysis}  shows that some of the improved performance may also be due to artifacts in \dataset{}.

We notice smaller improvements from XLNet QA over BERT QA ($4.12$ in $F_1$ test score, $2.6$ in EM test score) on \dataset{} compared to other reading comprehension benchmarks: SQuAD and RACE (see \citealp{Yang2019XLNetGA}). This might indicate the insufficiency of pretraining on more data (XLNet was pretrained on 6 times more plain text, nearly 10 times more wordpieces than BERT), for coreferential reasoning.

The passage-only baseline under-performs all other systems; examining its predictions reveals that it almost always predicts the most frequent entity in the passage. Its relatively low performance, despite the tendency for Wikipedia articles and passages to be written about a single entity, indicates that a large majority of questions likely require coreferential reasoning.

\subsection{Error Analysis}~\label{sec:error_analysis}
We analyzed the predictions from the baseline systems to estimate the extent to which they really understand coreference.

Since the contexts in \dataset{} come from Wikipedia articles, they are often either about a specific entity, or are narratives with a single protagonist. We found that the baseline models exploit this property to some extent. We observe that 51\% of the \dataset{} questions in the development set that were correctly answered by BERT QA were either the first or the most frequent entity in the paragraphs, while in the case of those that were incorrectly answered, this value is 12\%. XLNet QA also exhibits a similar trend, with the numbers being 48\% and 11\%, respectively.

QA systems trained on \dataset{} often need to find entities that occur far from the locations in the paragraph at which that the questions are anchored. To assess whether the baseline systems exploited answers being close, we manually analyzed predictions of BERT QA and XLNet QA on 100 questions in the development set, and found that the answers to 17\% of the questions correctly answered by XLNet QA are the nearest entities, whereas the number is 4\% for those incorrectly answered. For BERT QA, the numbers are 17\% and 6\% respectively.

%% file: tables/baselines_results.tex
\begin{table}[!t]
	\centering
	\resizebox{\columnwidth}{!}{
		\begin{tabular}{lrrrr}
			\toprule
			\multirow{2}{*}{\bf Method}    & \multicolumn{2}{c}{\bf Dev} & \multicolumn{2}{c}{\bf Test} \\
			\cmidrule(lr){2-3}
			\cmidrule(lr){4-5}
			& EM & $F_1$ & EM&  $F_1$\\
			 \midrule
			  \multicolumn{5}{l}{\textbf{Heuristic Baselines}}\\
			   passage-only BERT QA & 19.77 & 25.56 & 21.25 & 29.27 \\
			    passage-only XLNet QA & 18.44 & 27.59 & 18.57 & 28.24 \\
			    \addlinespace
			    \multicolumn{5}{l}{\textbf{Reading Comprehension}}\\
			     QANet & 34.41 & 38.26 & 34.17 & 38.90\\
			      QANet+BERT & 43.09 & 47.38 & 42.41 & 47.20\\
			       BERT QA & 58.44 & 64.95 & 59.28 & 66.39\\
			        XLNet QA & \textbf{64.52} & \textbf{71.49} & \textbf{61.88} & \textbf{70.51} \\
				\addlinespace
				\textbf{Human Performance} & - & - & 86.75 & 93.41 \\
				\bottomrule
		\end{tabular}
		}
		\caption{Performance of various baselines on \dataset{}, measured by exact match (EM) and $F_1$. Boldface marks the best systems for each metric and split.
		}
		\label{tab:baselines_results}
\end{table}

%% file: appendix.tex
\section{Crowdsourcing Logistics}\label{sec:crowdsourcing_logistics}
\subsection{Instructions}
The crowdworkers were giving the following instructions:

``In this task, you will look at paragraphs that contain several phrases that are references to names of people, places, or things. For example, in the first sentence from sample paragraph below, the references Unas and the ninth and final king of Fifth Dynasty refer to the same person, and Pyramid of Unas, Unas's pyramid and the pyramid refer to the same construction. You will notice that multiple phrases often refer to the same person, place, or thing. Your job is to write questions that you would ask a person to see if they understood that the phrases refer to the same entity. To help you write such questions, we provided some examples of good questions you can ask about such phrases. We also want you to avoid questions that can be answered correctly by someone without actually understanding the paragraph. To help you do so, we provided an AI system running in the background that will try to answer the questions you write. You can consider any question it can answer to be too easy. However, please note that the AI system incorrectly answering a question does not necessarily mean that it is good. Please read the examples below carefully to understand what kinds of questions we are interested in.''

\subsection{Examples of Good Questions}
We illustrate examples of good questions for the following paragraph.

\paragraph{}``The Pyramid of Unas is a smooth-sided pyramid built in the 24th century BC for the Egyptian pharaoh Unas, the ninth and final king of the Fifth Dynasty. It is the smallest Old Kingdom pyramid, but significant due to the discovery of Pyramid Texts, spells for the king's afterlife incised into the walls of its subterranean chambers. Inscribed for the first time in Unas's pyramid, the tradition of funerary texts carried on in the pyramids of subsequent rulers, through to the end of the Old Kingdom, and into the Middle Kingdom through the Coffin Texts which form the basis of the Book of the Dead. Unas built his pyramid between the complexes of Sekhemket and Djoser, in North Saqqara. Anchored to the valley temple via a nearby lake, a long causeway was constructed to provide access to the pyramid site. The causeway had elaborately decorated walls covered with a roof which had a slit in one section allowing light to enter illuminating the images. A long wadi was used as a pathway. The terrain was difficult to negotiate and contained old buildings and tomb superstructures. These were torn down and repurposed as underlay for the causeway. A significant stretch of Djoser's causeway was reused for embankments. Tombs that were on the path had their superstructures demolished and were paved over, preserving their decorations.''

\paragraph{}The following questions link pronouns:

\begin{enumerate}[label=\bfseries Q\arabic*:,leftmargin=*,labelindent=1em]
		    \item What is the name of the person whose pyramid was built in North Saqqara? \textbf{A:} Unas
			        \item What is significant due to the discovery of Pyramid Texts? \textbf{A:} The Pyramid of Unas
					    \item What were repurposed as underlay for the causeway? \textbf{A:} old buildings; tomb superstructures
\end{enumerate}

\paragraph{}The following questions link other references:

\begin{enumerate}[label=\bfseries Q\arabic*:,leftmargin=*,labelindent=1em]
		    \item What is the name of the king for whose afterlife spells were incised into the walls of the pyramid? \textbf{A:} Unas
			        \item Where did the final king of the Fifth dynasty build his pyramid? \textbf{A:} between the complexes of Sekhemket and Djoser, in North Saqqara
\end{enumerate}

\subsection{Examples of Bad Questions}
We illustrate examples of bad questions for the following paragraph.

\paragraph{} ``Decisions by Republican incumbent Peter Fitzgerald and his Democratic predecessor Carol Moseley Braun to not participate in the election resulted in wide-open Democratic and Republican primary contests involving fifteen candidates. In the March 2004 primary election, Barack Obama won in an unexpected landslide—which overnight made him a rising star within the national Democratic Party, started speculation about a presidential future, and led to the reissue of his memoir, Dreams from My Father. In July 2004, Obama delivered the keynote address at the 2004 Democratic National Convention, seen by 9.1 million viewers. His speech was well received and elevated his status within the Democratic Party. Obama's expected opponent in the general election, Republican primary winner Jack Ryan, withdrew from the race in June 2004. Six weeks later, Alan Keyes accepted the Republican nomination to replace Ryan. In the November 2004 general election, Obama won with 70 percent of the vote. Obama cosponsored the Secure America and Orderly Immigration Act. He introduced two initiatives that bore his name: Lugar–Obama, which expanded the Nunn–Lugar cooperative threat reduction concept to conventional weapons; and the Federal Funding Accountability and Transparency Act of 2006, which authorized the establishment of USAspending.gov, a web search engine on federal spending. On June 3, 2008, Senator Obama—along with three other senators: Tom Carper, Tom Coburn, and John McCain---introduced follow-up legislation: Strengthening Transparency and Accountability in Federal Spending Act of 2008.''

\paragraph{} The following questions do not require coreference resolution:

\begin{enumerate}[label=\bfseries Q\arabic*:,leftmargin=*,labelindent=1em]
		    \item Who withdrew from the race in June 2004? \textbf{A:} Jack Ryan
			        \item What Act sought to build on the Federal Funding Accountability and Transparency Act of 2006? \textbf{A:} Strengthening Transparency and Accountability in Federal Spending Act of 2008
\end{enumerate}

\paragraph{} The following question has ambiguous answers:

\begin{enumerate}[label=\bfseries Q\arabic*:,leftmargin=*,labelindent=1em]
		    \item Whose memoir was called Dreams from My Father? \textbf{A:} Barack Obama; Obama; Senator Obama
\end{enumerate}

\subsection{Worker Pool Management}
Beyond training workers with the detailed instructions shown above, we ensured that the questions are of high quality by selecting a good pool of 21 workers using a two-stage selection process, allowing only those workers who clearly understood the requirements of the task to produce the final set of questions. Both the qualification and final HITs had 4 paragraphs per HIT for paragraphs from movie plot summaries, and 5 per HIT for the other domains, from which the workers could choose. For each HIT, workers typically spent 20 minutes, were required to write 10 questions, and were paid US\$7. 

\section{Experimental Setup Details}\label{sec:experimental_setup_details}

Unless otherwise mentioned, we adopt the original published procedures and hyperparameters used for each baseline.

\paragraph{BERT QA and XLNet QA} 
We use uncased BERT, and cased XLNet, but lowercase our data while processing. We train our model with a batch size of 10, sequence length of 512 wordpieces, and a stride of 128. We use the AdamW optimizer, with a learning rate of $3^{-5}$. We train for 10 epochs, 
checkpointing the model 
after $19399$ steps. We report the performance of the checkpoint which is the best on the dev set. 

\paragraph{QANet} Durining training, we truncate paragraphs to 400 (word) tokens during training and questions to 50 tokens. During evaluation, we truncate paragraphs to 1000 tokens and questions to 100 tokens.

\paragraph{Passage-only baseline} 
We keep the HPs setup used for training BERT QA and XLNet QA and replace questions with empty strings. 

